\documentclass[11pt]{article}

\usepackage[preprint]{acl}

\usepackage{times}
\usepackage{latexsym}

\usepackage[T1]{fontenc}

\usepackage[utf8]{inputenc}

\usepackage{microtype}

\usepackage{inconsolata}

\usepackage{graphicx}

\usepackage{amsmath, amssymb}
\usepackage{booktabs,multirow}
\usepackage{subcaption}
\usepackage{tabularx}
\usepackage{makecell}
\usepackage{xcolor}
\usepackage{booktabs}

\usepackage{array}  

\newcommand{\method}{\textsc{EGMemory}}

\title{Propose, Verify, Commit: Evidence-Grounded Memory for Long-Horizon Multi-Actor Conversations}

\author{
 \textbf{Zihao Lu\textsuperscript{1}}\thanks{Equal contribution.},
 \textbf{Zhihang Yuan\textsuperscript{1}}\footnotemark[1],
 \textbf{Litu Ou\textsuperscript{2}},
 \textbf{Lei Shi\textsuperscript{1}}\thanks{Corresponding author.}
\\
\\
 \textsuperscript{1}Alibaba Cloud Computing, Hangzhou, China
 \textsuperscript{2}The University of Edinburgh, Edinburgh, UK
\\
\{qinian.lzh,yuanzhihang.yzh, juetian.sl\}@alibaba-inc.com
\\
litu.ou@ed.ac.uk
}

\begin{document}
\maketitle

\begin{abstract}
Memory enables language model agents to use information beyond their context windows, but maintaining and accessing this information reliably remains difficult.
Long-horizon conversational memory is especially challenging in \emph{multi-actor} settings, where relevant evidence is distributed across participants and contexts, and previously established information may later be revised.
We introduce \method, a \emph{training-free agentic memory} framework centered on \emph{evidence chains} that preserve message-level evidence and provenance while tracking active states and their revision histories.
At write time, adaptive memory inspection and an evidence-grounded \emph{propose--verify--commit} protocol govern state updates.
At read time, adaptive evidence navigation iteratively identifies the state and supporting evidence required for a query, using conversational structure to narrow the search space and lexical- semantic relevance to rank candidates.
The system operates through prompting and tool use without memory-specific policy training.
\method\ achieves \textbf{68.2\%} on GroupMemBench and \textbf{77.9\%} on EverMemBench, outperforming the strongest evaluated baselines by \textbf{22.7} and \textbf{21.4} percentage points, respectively.
It further reaches \textbf{73.6\%} on the dyadic LoCoMo benchmark, supporting its applicability beyond multi-actor conversations.
Code is available at this
\href{https://anonymous.4open.science/r/EGmemory-2E11/}
{anonymous GitHub repository}.
\end{abstract}

\section{Introduction}
\label{sec:introduction}

Large language model agents increasingly operate over interactions
that extend far beyond a single context window.
Persistent memory has therefore become a central component of
long-horizon agents, enabling them to retain past interactions,
recover relevant information, and maintain continuity across
sessions~\citep{maharana24locomo,di26longmemeval}.
Recent systems have improved how conversational histories are
constructed, updated, organized, and
accessed~\citep{chhikara2025mem0,wujiang25amem,
li2025memosmemoryosai}.
Much of this development has been studied through dyadic
conversations and sustained user--assistant interactions.
Extending memory to conversations involving multiple participants
requires handling additional ambiguity about who provided
information, which discussion it concerns, and how it changes
over time.

Long-horizon \emph{multi-actor} conversations distribute information
across participants, threads, and evolving contexts.
Similar statements may concern different discussions, while
messages about the same topic may either introduce complementary
information or revise an earlier decision.
Recent benchmarks expose these challenges directly.
GroupMemBench~\citep{yang2026groupmembench} emphasizes group dynamics,
speaker-grounded information, and audience-dependent language,
while EverMemBench~\citep{chuanrui26evermembench} evaluates
multi-party and multi-group collaboration with cross-context
reasoning and temporally evolving information.
Successful memory access must therefore determine \emph{whose}
evidence is relevant, \emph{which} conversational state it concerns,
and \emph{whether} that state remains active.

These requirements pose two connected challenges for memory
systems: maintaining information as conversations evolve and
deciding which evidence to access for a particular input.
Memory consolidation and structured histories help preserve
information and its changes~\citep{chhikara2025mem0,
banerjee26apexmem,zhengxuan26seem}.
However, a relevant memory may describe an earlier state,
another participant's position, or a different discussion.
The agent may therefore need to inspect additional evidence
before updating memory or answering a query.
Adaptive retrieval and tool-based memory control support
this process by allowing each access to build on earlier
observations~\citep{trivedi23interleaving,packer24memgpt}.
Approaches such as AgeMem and Memory-R1 learn memory-control
policies through reinforcement learning~\citep{yu26agenticmemory,
yan26memoryr1}, requiring additional memory-specific training.
We seek to combine explicit evidence and state histories
with adaptive memory access, using the foundation model's
existing capabilities without such training.

We introduce \method, a \emph{training-free agentic memory}
framework for long-horizon multi-actor conversations, centered
on \emph{evidence chains}.
Each evidence node preserves a message and its provenance, while an evidence chain tracks revisions of a particular conversational state and retains its supporting context.
This organization makes active information accessible alongside
the history it supersedes, preserving who contributed each
statement and how the state evolved.
Distinct states remain separate even when they share a topic,
while their evidence can be combined when a query spans
multiple discussions.
The resulting memory supports adaptive writing and reading
through the foundation model's existing tool-use capabilities,
without memory-specific policy training.

Our main contributions are as follows:
\begin{itemize}
    \item We organize long-horizon multi-actor memory around
    \textbf{evidence chains} that preserve message provenance
    and state history, distinguishing active information
    from the evidence it supersedes.

    \item We develop a \textbf{training-free agentic memory
    framework} with complementary writer and reader processes.
    Adaptive memory inspection and propose--verify--commit
    govern construction, while adaptive evidence navigation
    gathers support for downstream answers.

    \item We demonstrate substantial improvements on two
    long-horizon multi-actor benchmarks, exceeding the strongest
    evaluated baselines by \textbf{22.7} percentage points on
    GroupMemBench and \textbf{21.4} on EverMemBench, alongside
    a \textbf{4.3-point} improvement in the complementary
    dyadic LoCoMo evaluation~\citep{maharana24locomo}.
\end{itemize}

\section{Related Work}
\label{sec:related_work}

\paragraph{Conversational memory across participants.}
Long-term conversational memory requires retaining and applying
information across extended interactions.
LoCoMo evaluates memory over dyadic conversations~\citep{maharana24locomo},
while LongMemEval examines extraction, multi-session reasoning,
temporal reasoning, knowledge updates, and abstention in
user--assistant histories~\citep{di26longmemeval}.
Multi-actor settings introduce additional demands on attribution
and context.
GroupMemBench evaluates group dynamics, speaker-grounded beliefs,
and audience-adapted language~\citep{yang2026groupmembench},
while EverMemBench extends evaluation to multi-party, multi-group
conversations with interleaved topics and evolving
information~\citep{chuanrui26evermembench}.
These settings require memory to distinguish participants'
statements, preserve conversational context, and track changes
across discussions.

\paragraph{Memory construction and state evolution.}
Retrieval-based memory accesses stored histories through lexical
or dense matching, with hybrid search combining both signals.
The effectiveness of these operations also depends on how memory
is constructed.
MemoryBank summarizes interactions with selective forgetting and
reinforcement~\citep{wangjun24memorybank}, while SeCom studies memory
granularity and constructs topic-coherent
segments~\citep{zhuoshi25secom}.
Beyond storing passages or summaries, Mem0 reconciles extracted
information through addition, update, deletion, and no-op
operations~\citep{chhikara2025mem0}.
MemOS manages the lifecycle of plaintext, activation-based, and
parameter-level memories~\citep{li2025memosmemoryosai}.
Structured approaches further preserve relationships and history:
Zep maintains temporally evolving facts linked to episodic
sources~\citep{rasmussen2025zep};
APEX-MEM combines append-only temporal graphs with query-time
resolution of evolving information~\citep{banerjee26apexmem};
and SEEM links episodic event frames to source evidence for
context reconstruction~\citep{zhengxuan26seem}.
Together, these approaches extend memory construction from
selecting retrievable content to maintaining its organization,
provenance, and evolution.

\paragraph{Agentic memory management.}
Agentic memory gives language models control over memory
organization and access.
MemGPT exposes hierarchical memory management through
tools~\citep{packer24memgpt}, while A-MEM dynamically constructs
links and evolves interconnected memory
notes~\citep{wujiang25amem}.
Adaptive access also draws on multi-step retrieval:
IRCoT interleaves retrieval with reasoning so that earlier
observations guide subsequent searches~\citep{trivedi23interleaving},
and MAGMA uses policy-guided traversal over semantic, temporal,
causal, and entity graphs~\citep{jiang26magma}.
Other approaches learn memory-control behavior:
AgeMem and Memory-R1 use reinforcement learning to optimize
memory management and utilization~\citep{yu26agenticmemory,yan26memoryr1}.
This line of work treats memory operations as decisions
conditioned on the task and previously observed information.

Within this landscape, \method\ provides training-free
agentic memory for multi-actor conversations.
Evidence chains provide a shared representation for
preserving attributed state histories and supporting
adaptive writing and reading.


\section{Methodology}
\label{sec:method}

\begin{figure*}[t]
    \centering
    \includegraphics[width=\textwidth]{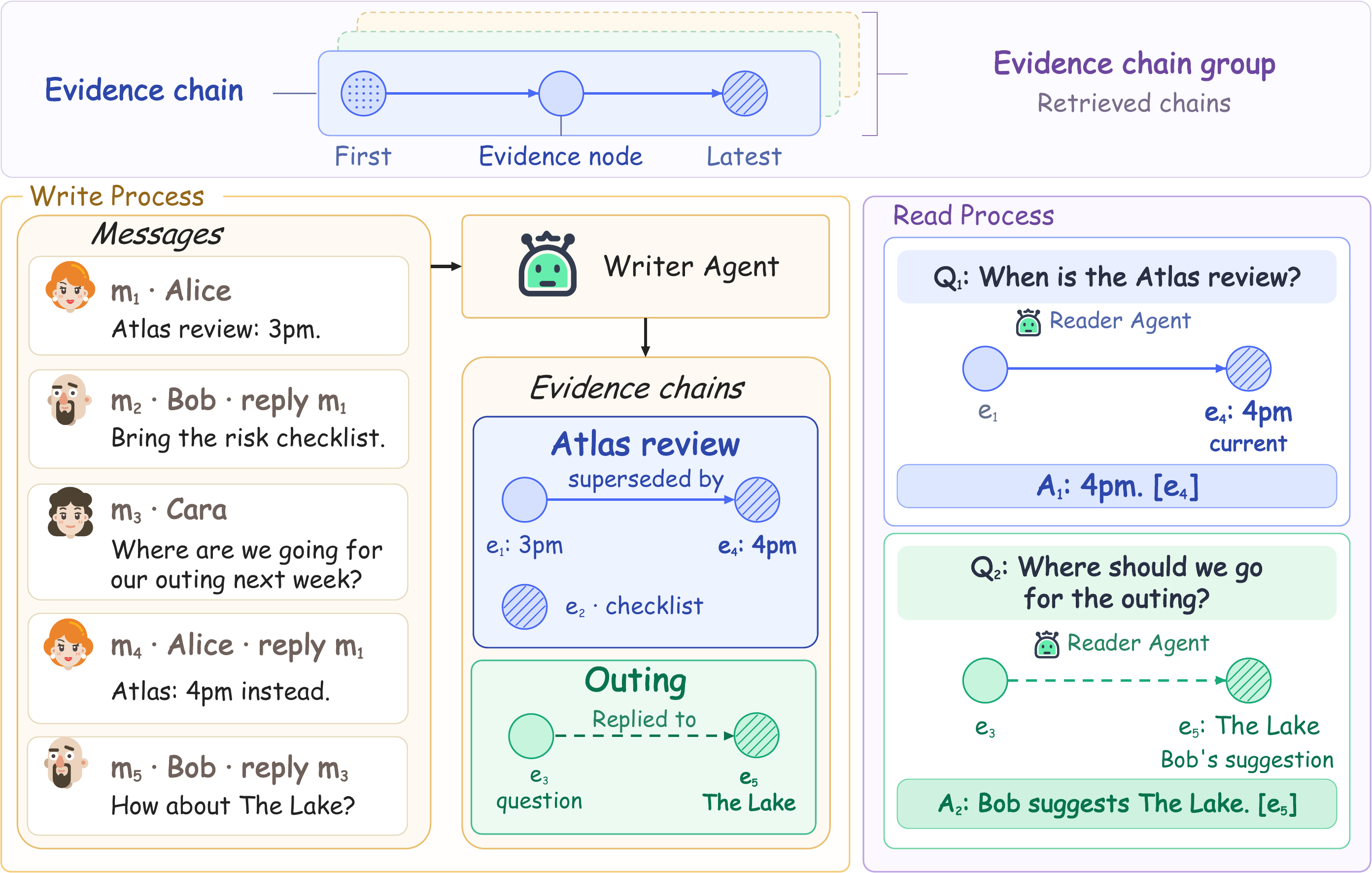}
    \caption{
        Overview of \method.
        \textbf{Top:} An evidence chain tracks an evolving
        conversational state. The legend marks the first node
        with dots; hatching denotes the latest node.
        Chains gathered during memory access form an
        \emph{evidence chain group}.
        \textbf{Left (Write):} The writer organizes five messages
        into three evidence chains: the Atlas review's time
        update, a separate singleton chain for the checklist,
        and the outing suggestion with its reply context.
        Solid arrows between evidence nodes denote supersession;
        dashed arrows denote reply relations.
        \textbf{Right (Read):} The reader answers with supporting
        evidence: 4\,pm supersedes 3\,pm for the review ($e_4$),
        while The Lake is attributed to Bob as a suggestion ($e_5$).
    }
    \label{fig:framework_overview}
\end{figure*}

\begin{table*}[t]
    \centering
    \small
    \setlength{\tabcolsep}{5pt}
    \renewcommand{\arraystretch}{1.08}
    \begin{tabularx}{\textwidth}{
        @{}l>{\raggedright\arraybackslash}X@{}
    }
        \toprule
        \textbf{Tool} & \textbf{Description} \\
        \midrule
        \multicolumn{2}{@{}l}{\textit{Writer agent}} \\
        \texttt{search\_registry}
        & Finds active states to associate incoming messages
          with existing memory. \\
        \texttt{propose\_update}
        & Submits a structured memory update for deterministic
          verification and commitment. \\
        \midrule
        \multicolumn{2}{@{}l}{\textit{Reader agent}} \\
        \texttt{search\_evidence}
        & Finds relevant evidence using two-pass hybrid search
          with structural scoping. \\
        \texttt{lookup\_phrase}
        & Finds exact phrases through literal-substring matching. \\
        \texttt{read\_current\_state}
        & Follows supersession links to return active states
          and supporting evidence. \\
        \texttt{resolve\_date}
        & Converts relative dates to absolute dates
          using a reference date. \\
        \bottomrule
    \end{tabularx}
    \caption{Tools for grounded memory construction and
    adaptive evidence navigation.}
    \label{tab:memory_tools}
\end{table*}

\subsection{Illustrative Overview}
\label{sec:framework_overview}

Figure~\ref{fig:framework_overview} illustrates \method\ through
two interleaved discussions.
Each \emph{evidence node} preserves a message and its provenance.
An \emph{evidence chain} tracks an evolving conversational state,
while chains assembled during memory access form an
\emph{evidence chain group}.

At write time, the writer organizes five messages into three
evidence chains.
The review-time chain links the initial 3\,pm time to its
revision to 4\,pm, preserving the superseded evidence.
The checklist reminder forms a separate singleton chain,
while the outing chain preserves Bob's suggestion with the
question as reply context.
At read time, the reader follows the relevant chains to answer
the review-time query with 4\,pm and the outing query with
Bob's suggestion of The Lake, preserving its attribution.

We next formalize the task, introduce the
\emph{propose--verify--commit} protocol for grounded memory
construction, and describe how evidence chains preserve
state updates and support adaptive reading.

\subsection{Problem Formulation}
\label{sec:problem_formulation}

We consider a long-horizon multi-actor conversation
$\mathcal{C}_t=(m_1,\ldots,m_t)$, where each message
$m_i=(x_i,a_i,\tau_i,c_i)$ contains its content, actor,
timestamp, and conversational context, respectively.
The writer incorporates incoming messages into memory,
while the reader answers queries whose supporting evidence
may span multiple actors and stages of the conversation.

We represent memory at step $t$ as
\begin{equation}
    \mathcal{M}_t =
    \left(\mathcal{E}_t,\mathcal{S}_t,\mathcal{R}_t\right),
    \label{eq:memory_state}
\end{equation}
where $\mathcal{E}_t$ contains evidence nodes with their
provenance, $\mathcal{R}_t$ records provenance and update
relations, including reply and supersession links, and
$\mathcal{S}_t$ identifies the nodes supporting active states.
Evidence persists,
$\mathcal{E}_{t-1}\subseteq\mathcal{E}_t$,
while active-state assignments may change as new messages arrive.

Both agents may need multiple memory accesses to resolve
the evidence relevant to an input $u$: an incoming message
for the writer or a query for the reader.
Starting from $H_0=\varnothing$, each access follows
\begin{equation}
\begin{aligned}
    r_j &= \pi(u,H_{j-1}), \\
    O_j &= \operatorname{Access}(\mathcal{M},r_j), \\
    H_j &= H_{j-1}\cup O_j ,
\end{aligned}
\label{eq:adaptive_access}
\end{equation}
where $\pi$ is the agent's access policy, $r_j$ is its request,
and $O_j$ contains the returned observations.
The accumulated observations $H_j$ guide subsequent requests
and support the writer's update or the reader's answer.

\subsection{Grounded Memory Construction}
\label{sec:memory_construction}

Given a new message $m_t$, the writer agent determines how
it relates to existing conversational states.
Because this relation may be implicit, the writer first uses
\texttt{search\_registry} to inspect the channel's active
states and reuse existing topics
(Table~\ref{tab:memory_tools}).
Following Eq.~\ref{eq:adaptive_access}, successive requests
can depend on earlier observations, accumulated in $H_t^{w}$.
This interaction resolves which state, if any, the message
concerns; sharing a topic alone does not establish a revision
relation.

The writer then submits an update through a
\textbf{propose--verify--commit} protocol.
The agent supplies the proposal, while deterministic code
performs verification and commitment.

\paragraph{Propose.}
The writer invokes \texttt{propose\_update} to submit
\begin{equation}
    \Delta_t =
    \operatorname{Propose}
    \left(m_t,H_t^{w}\right),
    \label{eq:propose}
\end{equation}
containing the topic, a verdict describing the message's
relation to existing state, a source quote expressing the
state-bearing value when present, a prior-state reference
when applicable, and a context-resolved rewrite for
subsequent memory access.
For example, Alice's ``4\,pm instead'' proposes a revision
of the review time, whereas the checklist reminder concerns
a separate state under the same topic.

\paragraph{Verify.}
The system checks the proposal against the source message
and existing memory:
\begin{equation}
    \widehat{\Delta}_t =
    \operatorname{Verify}
    \left(
        \Delta_t;
        m_t,\mathcal{M}_{t-1}
    \right).
    \label{eq:verify}
\end{equation}
A proposed state-bearing value must be grounded in the source
evidence, and a claimed revision must resolve to a valid
prior state.
Our implementation uses deterministic overlap-based checks
to ground proposed values and prior-state references.
These checks separate the model's semantic interpretation
from its authority to modify persistent state: proposing
an update does not itself authorize a state change.

\paragraph{Commit.}
The system converts the verified update into a canonical
evidence node $e_t$ and updates memory:
\begin{equation}
    \left(e_t,\mathcal{M}_t\right)
    =
    \operatorname{Commit}
    \left(
        \mathcal{M}_{t-1},
        \widehat{\Delta}_t
    \right).
    \label{eq:commit}
\end{equation}
The node preserves its source message, context-resolved
representation, and provenance, including actor, timestamp,
topic, and structural context.
Commitment updates the evidence store, active-state registry,
and associated relations according to the verified proposal.
Section~\ref{sec:state_management} describes how the resulting
nodes are organized into evidence chains.

\subsection{Evidence Chains}
\label{sec:state_management}

Grounded memory construction determines how incoming messages
are committed; evidence chains organize the resulting nodes
and preserve state history.
An evidence chain tracks a particular conversational state
and retains its supporting context, rather than collecting
all information about a topic.

In Figure~\ref{fig:framework_overview}, the review-time chain
links $e_1$ (3\,pm) to $e_4$ (4\,pm), whereas the checklist
reminder $e_2$ forms a separate singleton chain.
The two chains share the Atlas topic but concern distinct
states.
Supersession links represent state revisions, while reply
links retain conversational context.

A verified update may establish, restate, or revise a state.
For a committed revision, the new node explicitly supersedes
the previous supporting node and supports the updated active
value.
The previous node remains in $\mathcal{E}_t$, retaining the
evidence and provenance of the earlier state.

For a particular state, let $e^{(k)}\prec e^{(k+1)}$ denote
that $e^{(k+1)}$ supersedes $e^{(k)}$.
Repeated revisions within a chain induce the sequence
\begin{equation}
    e^{(1)} \prec e^{(2)} \prec \cdots \prec e^{(K)},
    \label{eq:version_trajectory}
\end{equation}
where all nodes remain available as historical evidence,
while $e^{(K)}$ supports the active state.
Current-state queries can inspect the active supporting node,
whereas historical queries can follow the supersession
relations to earlier values and their evidence.

Evidence chains provide the shared memory interface for the
writer and reader.
The writer accesses existing evidence and active states to
resolve an incoming message before proposing an update.
The reader uses the same nodes, state assignments, and
relations to gather support for an answer.
Relevant chains assembled during memory access constitute
an \emph{evidence chain group}, allowing evidence about
distinct states to be considered together without merging
their revision histories.

\subsection{Adaptive Evidence Navigation}
\label{sec:memory_retrieval}

At query time, the relevant conversational state and its
supporting evidence may not be identifiable from the query
alone.
The reader agent therefore navigates evidence chains
adaptively, using earlier observations to guide subsequent
memory requests.

Following Eq.~\ref{eq:adaptive_access}, let $H_q^{r}$ denote
the accumulated read-time observations, including evidence
nodes and tool outputs.
Each request depends on the query and the observations so far,
allowing the reader to refine an ambiguous search, inspect an
active state, or gather support from additional chains.
The reader uses four tools (Table~\ref{tab:memory_tools}).

The primary tool, \texttt{search\_evidence}, performs two-pass
hybrid dense--lexical search with metadata scoping.
The first pass searches globally within the specified scope
and selects a small set of high-relevance anchors.
These anchors induce a query-conditioned candidate pool from
topic buckets and provenance neighborhoods, including reply
and supersession relations.
This expansion exposes evidence within and across chains;
optional participant constraints further restrict its scope.
The second pass reapplies hybrid search within the resulting
pool to recover additional relevant evidence.

Conversational structure constrains \emph{where to search},
while content relevance ranks candidates; structural
proximity is not an independent relevance signal.

Three targeted tools complement this search.
\texttt{read\_current\_state} follows supersession relations
to return active states and their supporting evidence nodes.
\texttt{lookup\_phrase} performs literal-substring matching
without semantic search to recover exact wording.
\texttt{resolve\_date} deterministically converts relative
date expressions to absolute dates using a reference date.
All tools share the same interaction history, so one tool's
output can guide the reader's next request.
Navigation ends when sufficient grounded evidence has been
collected or the available memory cannot support a justified
answer.

The reader generates the final response from the accumulated
observations:
\begin{equation}
    \hat{y}
    =
    \operatorname{Answer}
    \left(q,H_q^{r}\right),
    \label{eq:answer}
\end{equation}
allowing evidence from successive accesses and distinct
chains to jointly support the answer.

\section{Experiments}
\label{sec:experiments}

\begin{table*}[t]
    \centering
    \small
    \setlength{\tabcolsep}{9pt}
    \renewcommand{\arraystretch}{1.08}
    \begin{tabular}{lccc}
        \toprule
        &
        \multicolumn{2}{c}{\textbf{Long-Horizon Multi-Actor Memory}}
        & \textbf{Dyadic Generalization} \\
        \cmidrule(lr){2-3}
        \cmidrule(lr){4-4}
        \textbf{Method}
        & \textbf{GroupMemBench}
        & \textbf{EverMemBench}
        & \textbf{LoCoMo} \\
        \midrule

        \multicolumn{4}{l}{\textit{One-shot retrieval controls}} \\
        Dense RAG
        & 35.8 & 27.8 & 39.4 \\
        BM25 RAG
        & 45.5 & 30.4 & 47.3 \\
        Hybrid RAG
        & 43.9 & 32.1 & 48.1 \\

        \midrule
        \multicolumn{4}{l}{\textit{Long-term memory systems}} \\
        MemoryBank (\textsc{MB-F})
        & 30.7 & 1.9 & 28.3 \\
        MemoryBank (\textsc{MB-NF})
        & 40.7 & 54.2 & 68.6 \\
        MemOS
        & 20.3 & 25.0 & 56.2 \\
        A-MEM
        & 40.9 & 56.5 & 63.5 \\
        Mem0
        & 42.9 & 51.3 & 69.3 \\

        \midrule
        \textbf{\method}
        & \textbf{68.2}
        & \textbf{77.9}
        & \textbf{73.6} \\
        \quad {\footnotesize 95\% CI}
        & {\footnotesize [64.8, 71.4]}
        & {\footnotesize [76.2, 79.5]}
        & {\footnotesize [71.6, 75.5]} \\

        \midrule
        $\Delta$ vs.\ best baseline
        & \textbf{+22.7}
        & \textbf{+21.4}
        & \textbf{+4.3} \\
        \bottomrule
    \end{tabular}
    \caption{
        Overall accuracy (\%) on GroupMemBench, EverMemBench, and LoCoMo.
        One-shot controls retrieve the top-10 messages using the same answering and judging models.
        Confidence intervals are 95\% Wilson intervals for \method;
        $\Delta$ is the absolute improvement over the strongest baseline.
    }
    \label{tab:overall_results}
\end{table*}

\begin{table*}[t]
    \centering
    \small
    \setlength{\tabcolsep}{7pt}
    \begin{tabular}{lrrrrrrr}
        \toprule
        \textbf{Category}
        & \textbf{$n$}
        & \textbf{MB-F}
        & \textbf{MB-NF}
        & \textbf{MemOS}
        & \textbf{A-MEM}
        & \textbf{Mem0}
        & \textbf{\method} \\
        \midrule
        Multi-hop
        & 182 & 19.2 & 33.5 & 3.8 & 36.3 & 35.2
        & \textbf{75.3} \\
        Temporal
        & 162 & 18.5 & 32.1 & 3.1 & 29.6 & 46.3
        & \textbf{80.2} \\
        Abstention
        & 139 & 90.6 & 87.1 & \textbf{96.4} & 83.5 & 92.8
        & 79.1 \\
        Knowledge update
        & 107 & 12.1 & 30.8 & 0.0 & 29.0 & 16.8
        & \textbf{46.7} \\
        Term ambiguity
        & 106 & 11.3 & 12.3 & 1.9 & 22.6 & 20.8
        & \textbf{45.3} \\
        User implicit
        & 49 & 26.5 & 46.9 & 6.1 & 40.8 & 24.5
        & \textbf{67.3} \\
        \bottomrule
    \end{tabular}
    \caption{
        Category-level accuracy (\%) on GroupMemBench.
        \textsc{MB-F} and \textsc{MB-NF} denote MemoryBank with
        forgetting enabled and disabled, respectively.
    }
    \label{tab:gmb_breakdown}
\end{table*}

\begin{table*}[t]
    \centering
    \small
    \setlength{\tabcolsep}{7pt}
    \begin{tabular}{lrrrrrrr}
        \toprule
        \textbf{Task}
        & \textbf{$n$}
        & \textbf{MB-F}
        & \textbf{MB-NF}
        & \textbf{MemOS}
        & \textbf{A-MEM}
        & \textbf{Mem0}
        & \textbf{\method} \\
        \midrule
        \multicolumn{8}{l}{\textit{Fine-Grained Recall}} \\
        \quad Single-hop Retrieval
        & 213 & 1.9 & 84.0 & 25.8 & 82.2 & 79.8
        & \textbf{97.2} \\
        \quad Multi-hop Trajectory
        & 249 & 0.0 & 2.4 & 0.4 & 3.6 & 1.6
        & \textbf{74.7} \\
        \quad Temporal Duration
        & 300 & 0.3 & 14.7 & 7.3 & 21.7 & 11.3
        & \textbf{45.0} \\

        \midrule
        \multicolumn{8}{l}{\textit{Memory Awareness}} \\
        \quad Constraint
        & 402 & 4.2 & 70.1 & 43.5 & 71.4 & 70.6
        & \textbf{92.8} \\
        \quad Proactivity
        & 427 & 1.2 & 77.0 & 32.6 & 80.6 & 73.3
        & \textbf{96.7} \\
        \quad Update
        & 268 & 3.0 & 86.9 & 35.1 & 88.1 & 88.8
        & \textbf{97.4} \\

        \midrule
        \multicolumn{8}{l}{\textit{Profile Understanding}} \\
        \quad Style
        & 176 & 0.0 & 51.1 & 17.0 & \textbf{52.3} & 36.4
        & 32.4 \\
        \quad Skill
        & 169 & 1.8 & 28.4 & 23.7 & 34.9 & 29.6
        & \textbf{58.0} \\
        \quad Role
        & 196 & 3.6 & 46.4 & 21.9 & 45.4 & 37.8
        & \textbf{71.4} \\
        \bottomrule
    \end{tabular}
    \caption{
        Category-level accuracy (\%) on EverMemBench, grouped by
        Fine-Grained Recall, Memory Awareness, and Profile Understanding.
    }
    \label{tab:evermem_breakdown}
\end{table*}

\begin{table*}[t]
    \centering
    \small
    \setlength{\tabcolsep}{7pt}
    \begin{tabular}{lrrrrrrr}
        \toprule
        \textbf{Category}
        & \textbf{$n$}
        & \textbf{MB-F}
        & \textbf{MB-NF}
        & \textbf{MemOS}
        & \textbf{A-MEM}
        & \textbf{Mem0}
        & \textbf{\method} \\
        \midrule
        Single-hop
        & 841 & 9.8 & 73.6 & 57.3 & 66.5 & 75.5
        & \textbf{81.8} \\
        Adversarial
        & 446 & \textbf{95.1} & 86.3 & 85.9 & 84.8 & 86.1
        & 85.0 \\
        Temporal
        & 321 & 5.3 & 67.0 & 38.9 & 60.4 & 70.1
        & \textbf{71.7} \\
        Multi-hop
        & 282 & 9.2 & 36.5 & 33.3 & 32.3 & 36.9
        & \textbf{43.6} \\
        Open-domain
        & 96 & 14.6 & 41.7 & 34.4 & 40.6 & 29.2
        & \textbf{43.8} \\
        \bottomrule
    \end{tabular}
    \caption{
        Category-level accuracy (\%) on LoCoMo, used as the dyadic
        transfer evaluation.
    }
    \label{tab:locomo_breakdown}
\end{table*}

\begin{table}[t]
    \centering
    \small
    \setlength{\tabcolsep}{9pt}
    \renewcommand{\arraystretch}{1.08}
    \begin{tabular}{lcc}
        \toprule
        \textbf{Configuration}
        & \textbf{Overall}
        & $\mathbf{\Delta}$ \\
        \midrule
        \textbf{\method}
        & \textbf{68.2}
        & -- \\

        \midrule
        \multicolumn{3}{l}{\textit{Reader-side}} \\
        w/o Reranker
        & 63.5 & $-4.7$ \\
        w/o Versioned state
        & 63.2 & $-5.0$ \\
        Single-round reader
        & 60.3 & $-7.9$ \\

        \midrule
        \multicolumn{3}{l}{\textit{Writer-side}} \\
        w/o Evidence verification
        & 65.9 & $-2.3$ \\
        \bottomrule
    \end{tabular}
    \caption{
        Ablation on GroupMemBench.
        $\Delta$ denotes the absolute accuracy change from the full system.
    }
    \label{tab:ablation}
\end{table}

\subsection{Overview}
\label{sec:exp_overview}

Our evaluation focuses primarily on long-horizon \emph{multi-actor}
memory. We use GroupMemBench~\citep{yang2026groupmembench} and
EverMemBench~\citep{chuanrui26evermembench} as the two primary
benchmarks, while LoCoMo~\citep{maharana24locomo} provides a
complementary test of generalization to conventional dyadic
conversations.

We evaluate the system at three levels. We first compare end-to-end
performance against one-shot retrieve--then--read controls and
representative long-term memory systems, separating the benefit of
persistent stateful memory from retrieval alone. We then examine
category-level performance across memory capabilities, and finally
ablate the write- and read-side mechanisms of \method.

\subsection{Experimental Setup}
\label{sec:exp_setup}

\paragraph{Datasets.}
GroupMemBench provides 120,000 messages for memory construction and
745 evaluation questions spanning Multi-hop, Temporal, Abstention,
Knowledge update, Term ambiguity, and User implicit reasoning,
directly stressing distributed evidence and evolving conversational state.
EverMemBench provides 51,023 messages and 2,400 evaluation questions
covering \textit{Fine-Grained Recall}, \textit{Memory Awareness},
and \textit{Profile Understanding}, providing complementary evaluation
of cross-context retrieval, updates, and participant characteristics.
We additionally evaluate on LoCoMo, using 5,882 messages for memory
construction and 1,986 questions spanning Single-hop, Adversarial,
Temporal, Multi-hop, and Open-domain reasoning; LoCoMo serves as a
generalization benchmark rather than a primary optimization target.

\paragraph{Baselines and retrieval controls.}
Our one-shot controls use BM25, dense retrieval with \texttt{text-embedding-v4}, or reciprocal-rank fusion of the two (Hybrid RAG), retrieving the top-10 messages before a single answering step. These controls test whether the gains of \method\ can be explained by conventional retrieval without iterative access or explicit state maintenance.
We additionally compare against Mem0~\citep{chhikara2025mem0}, A-MEM~\citep{wujiang25amem}, MemOS~\citep{li2025memosmemoryosai}, and MemoryBank~\citep{wangjun24memorybank}; for MemoryBank, we report variants with (\textsc{MB-F}) and without (\textsc{MB-NF}) forgetting.
All systems use the same benchmark-specific answering and judging
protocol.

\paragraph{Implementation.}
We use \texttt{qwen3.7-max} for memory construction and question
answering, \texttt{Kimi-K3} as the evaluation judge,
\texttt{text-embedding-v4} for dense retrieval, and
\texttt{qwen3-rerank} for reranking.
The writer and reader operate through prompting and tool calls without memory-specific training or fine-tuning.
The one-shot controls use the same answering model as \method, while \method\ additionally performs the adaptive write- and read-time interactions described in Section~\ref{sec:method}.
Supplementary methodological and experimental details appear
in Appendix~\ref{app:supplementary}.

\paragraph{Evaluation protocol.}
We report accuracy as the primary metric together with category-level results for each benchmark and 95\% Wilson confidence intervals for aggregate \method\ results.
Final answers from all systems are evaluated with \texttt{Kimi-K3} under the same benchmark-specific judging protocol.
Controlled component ablations are conducted on GroupMemBench and
reported as absolute percentage-point changes from the full system.

\subsection{Overall Performance}
\label{sec:overall_results}

Table~\ref{tab:overall_results} summarizes the overall results.
\method\ achieves the highest accuracy among the evaluated systems
on all three benchmarks.
It reaches \textbf{68.2\%} on GroupMemBench and \textbf{77.9\%}
on EverMemBench, exceeding the strongest evaluated baselines by
\textbf{22.7} and \textbf{21.4} percentage points, respectively.
On LoCoMo, \method\ reaches \textbf{73.6\%}
(95\% CI: 71.6--75.5), improving over the strongest evaluated
baseline by \textbf{4.3 percentage points} and supporting its
applicability to dyadic conversations.

The strongest one-shot retrieval controls reach 45.5\% on
GroupMemBench and 32.1\% on EverMemBench.
Hybrid search does not consistently improve over lexical
retrieval: it trails BM25 on GroupMemBench and exceeds it by
only 1.7 percentage points on EverMemBench.
The ablations below examine how access to state histories,
iterative reading, and relevance ranking contribute to this
performance.

\subsection{Performance across Memory Capabilities}
\label{sec:capability_results}

We next examine category-level performance to characterize how the overall gains are distributed across different memory capabilities.

\subsubsection{Multi-Actor Memory}
\label{sec:multi_actor_results}

\paragraph{GroupMemBench.}
Table~\ref{tab:gmb_breakdown} shows that \method\ achieves the best
performance in five of the six categories.
Relative to the strongest memory-system baselines, the gains are
\textbf{+39.0} on Multi-hop, \textbf{+33.9} on Temporal,
\textbf{+22.7} on Term ambiguity, \textbf{+20.4} on User implicit,
and \textbf{+15.9} on Knowledge update.
These improvements are largest on tasks requiring distributed,
context-dependent, or evolving evidence.
Abstention is the only exception.
This is consistent with a coverage--abstention trade-off: the reader is designed to continue searching for supporting evidence before refusing, making it less conservative on no-answer cases.

\paragraph{EverMemBench.}
Table~\ref{tab:evermem_breakdown} reports results across
\textit{Fine-Grained Recall}, \textit{Memory Awareness}, and
\textit{Profile Understanding}.
\method\ leads on eight of the nine tasks.
The largest gain occurs on Multi-hop Trajectory, where
\method\ achieves \textbf{74.7\%}, compared with 3.6\% for
the strongest evaluated baseline.
It also exceeds 90\% on all three Memory Awareness tasks.
These results show that the gains extend beyond factual recall
to reasoning over distributed and evolving conversational
information.
Style is the only exception, suggesting that distributed
stylistic preferences are less directly captured by the
current state-oriented memory representation.

\subsubsection{Generalization to Dyadic Conversations}
\label{sec:locomo_results}

We additionally evaluate \method\ on LoCoMo as a complementary test of
whether a memory architecture designed for multi-actor interactions
generalizes to conventional dyadic conversations.

As shown in Table~\ref{tab:locomo_breakdown}, \method\ achieves the
best performance on Single-hop, Temporal, Multi-hop, and Open-domain
questions.
Relative to the strongest memory-system baselines, it improves
Single-hop by \textbf{6.3} points, Multi-hop by \textbf{6.7},
Temporal by \textbf{1.6}, and Open-domain by \textbf{2.1}.
Adversarial questions remain the only category where \method\ does not
lead.
Together with the overall LoCoMo result, these findings indicate that
the proposed architecture generalizes beyond the multi-actor setting
without being specifically optimized for dyadic memory.

\subsection{Ablation Study}
\label{sec:ablation}

\paragraph{Read-time evidence resolution.}
Restricting the reader to a single interaction round produces
the largest accuracy drop (\textbf{7.9} percentage points),
followed by removing access to versioned state
(\textbf{5.0} points) and cross-encoder reranking
(\textbf{4.7} points).
The single-round reader retains the same tools and retrieval
pipeline, so its deficit measures the benefit of adapting later
requests to earlier observations.
Nevertheless, it achieves 60.3\% accuracy, above the strongest
one-shot retrieval control at 45.5\%.
The full system's advantage therefore extends beyond repeated
access: substantial gains remain with a single reader round,
while iterative navigation provides a further improvement.

\paragraph{Write-time state control.}
Removing evidence grounding for proposed state values and prior-state
references lowers accuracy by \textbf{2.3 points}.
Its smaller effect is consistent with its role as a state-integrity
gate: verification controls whether model-proposed interpretations may
modify persistent state rather than directly improving retrieval
ranking.

\section{Conclusion}
\label{sec:conclusion}

We studied long-horizon memory for multi-actor conversations, where
evidence is distributed across participants and contexts while
previously established information may continue to evolve.
We introduced \method, a \emph{training-free agentic memory}
framework centered on \emph{evidence chains} that preserve
message-level evidence and provenance while tracking active
states and their revision histories.

\method\ comprises two complementary processes.
At write time, adaptive memory inspection and an evidence-grounded
propose--verify--commit protocol govern state updates.
At read time, adaptive evidence navigation identifies the relevant
state and supporting evidence through iterative memory access.
Both processes operate over the same evidence chains using a
tool-capable foundation model, without memory-specific policy
training.

Across the two primary multi-actor benchmarks, \method\ outperforms
the strongest evaluated baselines by \textbf{22.7} percentage points
on GroupMemBench and \textbf{21.4} percentage points on EverMemBench,
while reaching \textbf{73.6\%} on the dyadic LoCoMo benchmark.
These results support the value of organizing conversational memory
around attributed evidence and state histories that agents can
adaptively access as interactions evolve.

\section*{Limitations}

\method\ has several limitations.
First, deterministic verification relies on source-grounding heuristics
rather than full semantic entailment, so strongly paraphrased revisions
may be rejected while lexical overlap alone cannot guarantee semantic
correctness.
Second, the explicit-state abstraction is better suited to factual and
decision-like information than to diffuse behavioral signals such as
style, consistent with the remaining weakness on EverMemBench.
Third, adaptive write- and read-time interaction requires more model
calls than one-shot retrieval, and this work does not optimize latency
or token cost.
Our evaluation also uses a single foundation-model family and
English-language benchmarks, leaving cross-model, multilingual, and
noisy speaker-attribution settings for future work.
Finally, persistent multi-party memory can contain sensitive
information; practical deployments should pair retention with
appropriate access-control, deletion, and privacy policies.



\bibliography{custom}

\appendix

\appendix

\section{Supplementary Details}
\label{app:supplementary}

This appendix supplements the methodology and experimental
protocol. The anonymous repository linked in the abstract
provides the implementation, writer and reader prompts,
configuration settings, and benchmark execution code.

\subsection{Supplementary Method Details}
\label{app:method_details}

\paragraph{Evidence representation.}
Each message-level evidence node retains its source content,
context-resolved rewrite, and provenance.
Dense search, lexical search, reranking, and passage presentation
use the rewritten text, falling back to the source content
when no rewrite is available.
Returned passages include available metadata such as speaker,
channel, topic, timestamp, and reply context.
Source messages remain stored for grounding and reconstruction.

\paragraph{State updates and histories.}
The writer inspects a channel-specific registry before proposing
an update.
Deterministic checks ground proposed values and references
in the source message and associate revisions with existing
registry entries.
At read time, revision links must additionally satisfy the
grounding criterion for authoritative supersession; weaker
links do not establish that an earlier value is obsolete.
Historical nodes remain available.
The implementation also provides derived chain views that
combine successive versions for memory access while retaining
their underlying message-level evidence.

\paragraph{Reader observations.}
Search results are selected by relevance and then presented
chronologically.
Previously returned passages are suppressed in subsequent
searches for the same question.
Current-state inspection returns chain-specific candidates
and supporting evidence: sharing a topic does not imply that
one candidate supersedes another.
The reader may continue accessing memory until it answers or
exhausts its interaction budget, after which it generates a
final response from the accumulated observations.

\subsection{Supplementary Experimental Protocol}
\label{app:experimental_protocol}

\paragraph{Models and evaluation.}
The writer and reader use \texttt{qwen3.7-max}, with
\texttt{text-embedding-v4} for dense representations and
\texttt{qwen3-rerank} for reranking.
The reported experiments use Kimi-K3 as the evaluation judge.
Writer and reader behavior is specified through prompts and
tools, without memory-specific training.
The full reader permits up to ten interaction rounds.
Answers are evaluated against reference answers under the
benchmark-specific judging protocol.
Failed evaluations are retained and counted as incorrect.
Aggregate confidence intervals use the Wilson method.

\paragraph{One-shot retrieval controls.}
BM25 RAG, Dense RAG, and Hybrid RAG each retrieve ten original
messages and pass them to the same answering model used by
\method.
Hybrid RAG combines lexical and dense rankings through
reciprocal-rank fusion.
Retrieved messages retain their available speaker and
conversational metadata.
These controls use neither iterative memory access nor
explicit state maintenance.
They compare conventional retrieval pipelines with the full
framework; the component ablations below examine individual
design choices.

\subsection{Ablation Definitions}
\label{app:ablation_details}

The ablations in Table~\ref{tab:ablation} modify the following
components while retaining the remaining experimental setup.

\paragraph{Without reranking.}
Cross-encoder reranking is disabled, leaving the fused
dense--lexical ranking.
Structural expansion, evidence chains, and adaptive reader
interaction remain available.

\paragraph{Without versioned state.}
Memory is constructed normally, but supersession relations
and derived chain views are excluded from read-time access.
Underlying message-level evidence, reply structure,
reranking, and adaptive reader interaction remain available.
This ablation tests access to state histories rather than
removing the original evidence.

\paragraph{Single-round reader.}
The reader is limited to one tool-enabled model turn,
which may issue multiple tool calls, followed by a final
answer without further tool access.
The tools and retrieval pipeline remain unchanged.
This removes subsequent requests conditioned on earlier
tool observations.

\paragraph{Without evidence verification.}
Source-grounding checks for proposed state values and
prior-state references are disabled.
The remaining verdict, state-association, and commitment
constraints are retained.
This ablation therefore removes the grounding gate rather
than the complete state-update procedure.

\end{document}